# Brain tumour identification using improved YOLOv8

Rupesh Dulal

Faculty of Engineering,
Tribhuvan University,
Kathmandu, Nepal

Rabin Dulal*

School of Computing, Mathematics and Engineering,
Charles Sturt University,
Wagga Wagga, Australia
Email: rdulal@csu.edu.au
*Corresponding author

**Abstract:** Identifying the extent of brain tumours is a significant challenge in brain cancer treatment. The main difficulty is in the approximate detection of tumour size. Magnetic resonance imaging (MRI) has become a critical diagnostic tool. However, manually detecting the boundaries of brain tumours from MRI scans is a labour-intensive task that requires extensive expertise. Deep learning and computer-aided detection techniques have led to notable advances in machine learning for this purpose. In this paper, we propose a modified you only look once (YOLOv8) model to detect the tumours within the MRI images accurately. The proposed model replaced the non-maximum suppression (NMS) algorithm with a real-time detection transformer (RT-DETR) in the detection head. NMS filters out redundant or overlapping bounding boxes in the detected tumours, but they are hand-designed and pre-set. RT-DETR removes hand-designed components. The second improvement was made by replacing the regular convolution block with ghost convolution. Ghost convolution reduces computational and memory costs while maintaining high accuracy and enabling faster inference, making it ideal for resource-constrained environments and real-time applications. The third improvement was made by introducing a vision transformer block in the backbone of YOLOv8 to extract context-aware features. We used a publicly available dataset of brain tumours in the proposed model. The proposed model performed better than the original YOLOv8 model and also performed better than other object detectors (faster R-CNN, mask R-CNN, YOLO, YOLOv3, YOLOv4, YOLOv5, SSD, RetinaNet, EfficientDet, and DETR). The proposed model achieved 0.91 mean average precision (mAP)@0.5.

**Biographical notes:** Rupesh Dulal is a Bachelor of Computer Engineering student at Tribhuvan University with a strong passion for software engineering and artificial intelligence. He is enthusiastic about building intelligent systems and enjoys exploring innovative solutions to real-world problems through coding and research. With a keen interest in emerging technologies, he actively engages in projects and learning opportunities that enhance his skills in AI and software development.

Rabin Dulal holds a Bachelor of Engineering in IT from Pokhara University (2016) and a Master of IT in Software Design and Development from Charles Sturt University (2019). He is currently completing his PhD in Computer Vision at Charles Sturt University. Rabin has professional experience as a software developer, AI engineer, and researcher. He has presented and published papers in various conferences and journals. His interests include computer vision, artificial intelligence, and software engineering. Rabin is a member of the Australian Computer Society (ACS), Nepal Engineers' Association (NEA), and the Institute of Electrical and Electronics Engineers (IEEE).

## 1 Introduction

A brain tumour represents a dangerous disease resulting from abnormal and unwanted cell growth in the brain. A brain tumour is the cause of death for thousands of people around the world every year (Bechet et al., 2014). Medical experts use brain MRI imaging technology to detect portions of tumours. This is the best approach used in the medical diagnosis system. However, detecting tumours by analysing the MRI images is labour-intensive, time-consuming, and can only be evaluated by the experts (Pedada et al., 2023). Therefore, an automatic and more straightforward solution is required to ease the process of brain tumour detection.

Brain tumours are of two types. One is malignant, and the other is non-malignant (Fossel et al., 1986). Malignant tumours can grow uncontrollably, invade surrounding tissues, and spread to other body parts. Malignant tumours are typically dangerous and require prompt treatment to prevent further spread and damage (Tamimi et al., 2015). Non-malignant tumours, also known as benign tumours, are not cancerous. They generally grow slowly and do not spread to other body parts. At the same time, benign tumours can still cause problems depending on their size and location (Tamimi et al., 2015). A primary brain tumour originates in the brain or spinal cord and starts to grow within these areas (The University of Kansas Cancer Center, 2024). A type of brain tumour known as glioma is responsible for most primary brain and spinal cord cancers in adults. On the other hand, meningiomas represent the majority of non-cancerous tumours. However, the situation is complicated because some meningiomas can act similarly to malignant tumours. At the same time, certain gliomas can be treated successfully and may stay in remission for many years or even be fully cured (The University of Kansas Cancer Center, 2024). Malignant tumours are more damaging to humans and challenging to detect (Badran et al., 2010). The primary purpose of this research is to help detect these two types of tumours from MRI photos using advanced technology.

In recent years, the advancement of artificial intelligence has shown impressive results in detecting and classifying brain tumours (Dahab et al., 2012; Pedada et al., 2023; Selvy et al., 2019; Maharjan et al., 2020). Various conventional machine-learning techniques are used to detect and classify tumours (Giraddi and Vaishnavi, 2017; Gurbinˇa et al., 2019; Halder et al., 2014). However, traditional machine learning techniques cannot automatically extract the features from the raw data. They need various hand-designed feature extraction techniques that require computer experts' knowledge (LeCun et al., 2015).

Convolutional neural network (CNN) is a deep learning model mostly used for processing data with a grid-like topology, like images. It is designed to automatically process the images as input and learn the features like textures, edges, and patterns. This process involves convolution layers and filters to extract or learn those local features. These patterns are combined and abstracted through deeper layers to recognise more complex structures, such as shapes or objects. CNNs are a compelling deep-learning method used for image recognition tasks (LeCun et al., 2015; Li et al., 2021; O'Shea and Nash, 2015; Gu et al., 2018; Yamashita et al., 2018). CNNs are very popular and widely used deep learning methods for detecting, identifying, and classifying brain tumours (Maharjan et al., 2020; Pedada et al., 2023; Selvy et al., 2019,?; Chattopadhyay and Maitra, 2022; Bathe et al., 2021; Bagyaraj et al., 2021; Kumar et al., 2021; Waghmare and Kolekar, 2021).

Despite the powerful capabilities of CNNs in processing spatial data, they struggle with tasks that require handling long sequences and capturing long-range dependencies. This limitation arises because CNNs are not inherently designed to maintain context over very long sequences where understanding relationships across distant elements is crucial (Vaswani et al., 2017). To address this gap, transformer networks were developed. Transformers utilise self-attention mechanisms to effectively capture dependencies across long sequences and enable context-aware processing (Vaswani et al., 2017).

The vision transformer (ViT) is a transformer-based network for image recognition (Dosovitskiy et al., 2020). Unlike traditional CNNs, ViT treats an image as a sequence of patches, enabling it to capture long-range dependencies and global context effectively. By leveraging self-attention mechanisms, the ViT excels in tasks like image classification and object detection, offering a powerful alternative to conventional convolutional approaches (Khan et al., 2022b; Park and Kim, 2022). Thus, ViT has shown impressive applications in many brain tumour classifications because of its powerful and context-aware feature extraction capacity (Tummala et al., 2022; Şahin et al., 2024; Hong et al., 2024; Khaniki et al., 2024).

The deep learning-based detection method is more powerful and beneficial. However, deep learning-based methods are more computationally expensive, which provides a challenge for resource management and fast detection of tumours. This research aims to provide a better solution with a balance of accuracy and speed in detecting tumours. This research presents the following contributions:

1 This paper proposed a modified YOLOv8 (Ultralytics, n.d.) model for detecting brain tumours, incorporating several advanced components to enhance performance. First, we integrated a ViT (Dosovitskiy et al., 2020) as a context-aware feature extraction block. ViT block helps to capture long-range dependencies within the input image features. Next, we utilised the RT-DETR (Zhao et al., 2024) component to process these extracted features, employing an NMS-free detection head that enhances detection accuracy and efficiency. Additionally, we included ghost convolution (Han et al., 2020) in our design, which provides a lighter convolution operation, reducing computational complexity without compromising the model's effectiveness. The parameters of the YOLOv8 model are 11.2 million, while the proposed model is 9.04 million. The reduction in the parameter reduces computational complexity. Together, these modifications aim to improve the accuracy and efficiency of brain tumour detection.

2 We used a publicly available dataset and conducted extensive experiments in different object detection networks [faster R-CNN (Ren et al., 2015), mask R-CNN (He et al., 2017), YOLO (Redmon and Farhadi, 2017), YOLOv3 (Farhadi and Redmon, 2018), YOLOv4 (Bochkovskiy et al., 2020), YOLOv5 (Jocher, 2020), YOLOv8 (Ultralytics, n.d.), SSD (Li and Zhou, 2017), RetinaNet (Lin et al., 2017b), EfficientDet (Tan et al., 2020), and DETR (Carion et al., 2020)] and compared the results with the proposed model. Our model outperformed the other object detection models.

## 2 Background terms

This section provides a brief introduction and background knowledge to understand the different object detection models and terms used in this research.

*Faster R-CNN* uses a fully convolutional region proposal network (RPN). It uses an input image of any size and produces a set of regional proposals and objectiveness scores. Moreover, it uses multi-scale anchor boxes to classify and localise objects in different scales and aspect ratios. Anchor boxes are predefined bounding boxes that help the model handle multiple objects of different shapes and sizes more efficiently, improving detection performance. Using multi-scale anchor boxes and fully convolutional networks allows faster R-CNN to accurately and efficiently detect objects across various scales and aspect ratios within an image.

*Mask R-CNN* extends faster R-CNN to deal with instance segmentation. The architecture of mask R-CNN is composed of feature pyramid networks (FPN) (He et al., 2017) with CNNs to extract hierarchical features (features extracted from the different levels of the deep networks having various sizes and resolutions). As we go deeper into the deep networks, the resolution of the feature maps decreases. FPN combines lower and higher-resolution feature maps in the network. In the first stage, RPN selects all feature maps generated by FPN and CNN (backbone) and generates region proposals. It also binds feature maps with locations in the original image using anchor boxes. The second stage uses another neural network to classify and localise the image.

Unlike faster R-CNN and mask R-CNN, whose operation is performed in two stages (the first stage for the generation of region proposals and the second stage is for the detection and identification), some models perform both tasks in a single stage. These models are called one-stage object detection models. *SSD* is a popular single-stage model. SSD performs object identification at one time or in a single shot. In SSD a single network is responsible for extracting the features from the image, localise the object by providing appropriate bounding boxes, and classify the objects by providing class names. SSD achieves efficiency by applying convolutional filters to feature maps at different scales to simultaneously detect objects at multiple resolutions. This method allows SSD to be fast and suitable for real-time object detection tasks.

*RetinaNet* is also a popular single-stag object detection model. RetinaNet introduces a novel focal loss function (Lin et al., 2017b) that down-weights the loss assigned to well-classified examples and focuses more on complex, misclassified examples during training. This focal loss helps RetinaNet perform better, especially when detecting objects at various scales and dealing with imbalanced datasets.

*EfficientDet* is also a single-stage object detection model. It belongs to a family of models that aims to balance accuracy and efficiency by leveraging the principles of model scaling and compound scaling. EfficientDet builds upon the EfficientNet (Tan and Le, 2019) architecture, which optimises model depth, width, and resolution based on a compound coefficient. EfficientDet uses EfficientNet as a backbone network, and bidirectional feature pyramid network (BiFPN) (Tan et al., 2020) in the head network to enhance multi-scale feature fusion and better detection accuracy. BiFPN is an advanced FPN designed to improve the efficiency and effectiveness of multi-scale feature fusion in object detection models. Its bidirectional and weighted fusion approach makes it a powerful component for achieving high performance in object detection (Tan et al., 2020). EfficientDet achieves state-of-the-art performance with significantly fewer parameters than other models. This makes EfficientDet well-suited for resource-constrained environments or applications requiring real-time inference.

*YOLO* approaches the object detection problem as a single regression problem, where the network directly predicts bounding boxes and class probabilities from an entire image in one evaluation using CNN. This differs from traditional methods involving generating region proposals and performing classification separately. The algorithm divides the input image into a grid of cells, and each cell is responsible for predicting multiple bounding boxes and their associated confidence scores. YOLO employs intersection over union (IoU) to measure the accuracy of predicted bounding boxes against ground truth and uses non-maximum suppression to select the most accurate bounding box if multiple are generated for the same object. Each bounding box prediction includes parameters for the box's centre (x, y) relative to the grid cell, its width (w) and height (h) relative to the entire image, and a confidence score representing the IoU with the ground truth object. This approach allows YOLO to detect objects in real time with a single pass through the network. *YOLOv3* is an improvement over YOLO, designed to be faster and more accurate. It introduces several key enhancements, including FPN, to extract features at different scales, enabling the detection of objects of varying sizes. It uses a deeper type of CNN called Darknet-53 (Farhadi and Redmon, 2018) backbone network for better feature extraction, enhancing the model's ability to capture complex image patterns. Moreover, it predicts bounding boxes and class probabilities simultaneously at three different scales to detect objects of various sizes more effectively (in the head network of YOLOv3). *YOLOv4* further improves its backbone by introducing cross stage partial (CSP) (Wang et al., 2020) on Darknet53. CSP connection is a design strategy used in deep networks to improve learning efficiency and reduce computational complexity. Moreover, it incorporates various optimisation techniques, such as bag of freebies (BoF), and architectural modifications, such as bag of specials (BoS), to enhance performance. *YOLOv5* utilises advanced training strategies such as mosaic data augmentation and self-ensembling for improved performance and enhanced post-processing techniques for more accurate bounding box predictions. Mosaic data augmentation is a powerful technique that enhances training datasets by combining four same or different images into one, thereby increasing variation and improving the model's ability to generalise and detect objects under various conditions (Jocher, n.d.). *YOLOv8* introduces several enhancements over its predecessors, including more efficient architecture and improved training strategies, which lead to better performance in real-time object detection tasks. It maintains the core principle of YOLO by predicting bounding boxes and class probabilities directly from images in a single pass, making it highly effective for applications requiring fast and accurate detection. *DETR* has demonstrated impressive performance on object detection benchmarks, showcasing the potential of transformer-based architectures for complex computer vision tasks. By leveraging the transformer's ability to model global context and dependencies, DETR offers a novel approach to object detection that eliminates the need for many design choices and hyperparameters associated with traditional anchor-based methods. DETR is free of hand-designed components like anchor boxes and non-maximum suppression (NMS) and directly predicts the object class and bounding box. It uses a CNN backbone (pre-trained ResNet50) and transformer architecture in the head to detect the objects.

*RT-DETR* is a popular real-time object detection model which is fast in operation. It is motivated by DETR, meaning it eliminates the use of hand-designed anchor boxes and NMS for locating the detected or identified object. RT-DETR employs a hybrid encoder, which integrates convolutional and transformer-based layers to enhance efficiency and speed. The convolutional layers extract local features efficiently, while the transformer layers capture long-range dependencies and global context. This combination optimises processing by leveraging the strengths of both architectures – convolutions for fast, localised feature extraction and transformers for improved contextual understanding. As a result, RT-DETR achieves real-time performance without compromising accuracy. The model effectively handles multiscale features by separating intra-scale interactions from cross-scale fusion. Additionally, RT-DETR is highly versatile, allowing for flexible adjustments in inference speed by modifying decoder layers without requiring retraining. This is an improved version of DETR, which minimises the significant computational complexity in DETR.

*GhostConv* is a component used in the architecture of GhostNet (Han et al., 2020), which is a type of convolutional neural network (CNN) designed to be more efficient in terms of computational cost and model size. Feature maps obtained from Conv often contain significant redundancy, with many feature maps being similar. It is inefficient to rely solely on expensive convolution operations to generate these redundant feature maps. The GhostConv addresses this inefficiency using cheaper linear operations (ghost bottlenecks) to extract feature maps after the initial convolution. This approach allows the GhostConv block to achieve more functional outcomes than traditional convolution operations but with fewer parameters and lower computational costs. The ghost module enhances the network's overall efficiency without compromising performance by reducing redundancy and focusing computational resources more efficiently.

*Transformer encoder (TE)* is the building block of ViT. ViT consists of many TEs stacked together. A sketch of TE is shown in Figure 1. TE uses embedded patches as input and processes using its major components, multi-head attention, dropout, and multilayer perception (MLP).

**Figure 1** The architecture of the transformer encoder showing its major components (see online version for colours)

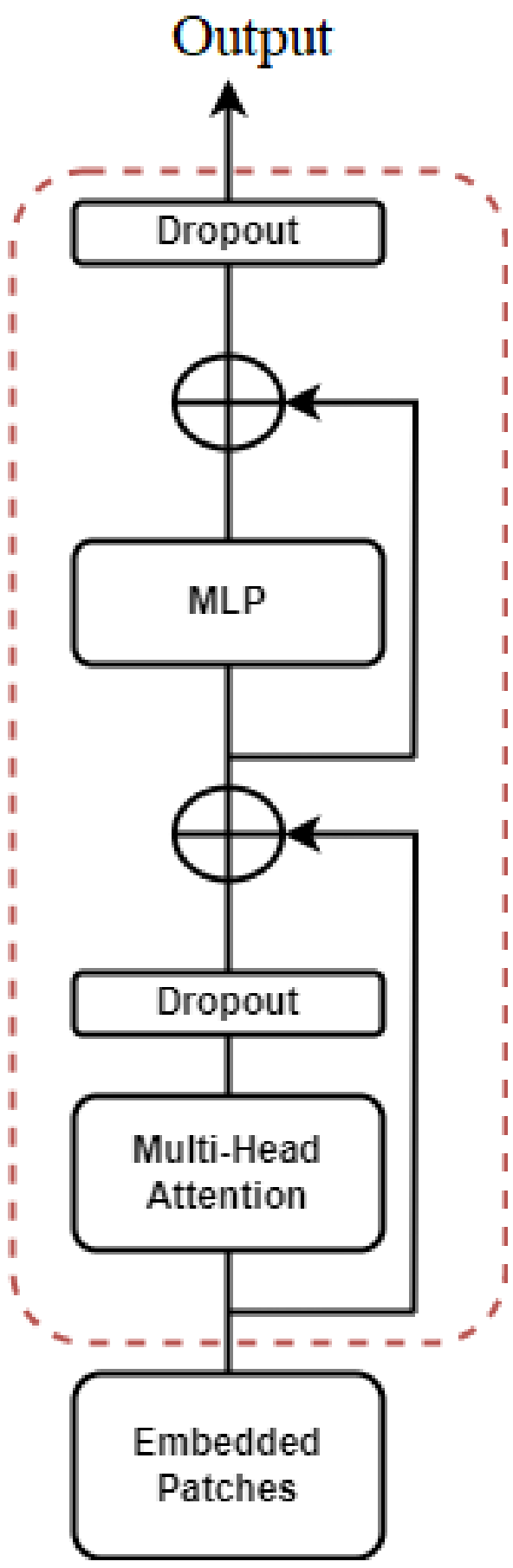

TE takes an input image. TE first divides an image into non-overlapping patches of the exact sizes. Each patch of the image is converted into a 1-dimensional vector called patch embedding. To make the sequence of the patches meaningful and in a fixed order, positional encoding is applied to each vector resulting from each patch. Positional encoding is an extra information added to maintain the position of each patch in the sequence of many input patches. The patches with positional encoding are called embedded patches. An alternate sine and cosine functions are used to calculate positional encoding. The below formula is used to generate positional encoding.

$$PE_{(pos,2i)} = \sin(pos/10{,}000^{2i/d_{model}}) \quad (1)$$

$$PE_{(pos,2i+1)} = \cos(pos/10{,}000^{2i/d_{model}}) \quad (2)$$

where

- $d_{\text{model}}$ – the dimension of the embeddings in the TE. For example, if the embeddings are 512-dimensional vectors, then $d_{\text{model}} = 512$.

- PE – positional encoding value for a given position $pos$ and dimension index $2i$ or $2i + 1$.
- $pos$ – the token's position in the sequence (e.g., the 1st, 2nd, 3rd token, etc.).
- $i$ – the index of the dimension of the positional encoding vector, running from $0$ to $\frac{d_{\text{model}}}{2} - 1$.

The multi-head attention block has three unique parameters: the query ($Q$), key ($K$), and value ($V$). In simple words, the query is the information to be processed, the key is the relevance of the information, and the value is the summary of the query and its relevance to the different individual components within the input sequence. The $W_i^Q$, $W_i^K$, and $W_i^V$ are learnable weight matrices corresponding to query, key, and value $i$. These parameters are initially assigned with random values and then updated during the training. For an input sequence vector $X$,

$$Q, K, V = XW^Q, XW^K, XW^V$$

$Q$, $K$, and $V$ are the matrix multiplication of the input sequence vector and the respective learnable parameters. The following equations obtain the outputs of the attention weights.

$$\text{Attention}(Q, K, V) = \text{softmax}\left(\frac{QK^T}{\sqrt{d_k}}\right)V \quad (3)$$

The attention scores of $Q$, $K$, and $V$ are scaled by $\frac{1}{\sqrt{d_k}}$ before applying the softmax function where $d_k$ is the dimension of the key. This is the attention of one head. The final output of the multi-head attention (MHA) block is obtained by the concatenation of the attention scores of each head through a linear transformation as obtained by the following equation:

$$\text{MHA}(Q, K, V) = \text{Concat}(head_1, \ldots, head_h) \cdot W_O \quad (4)$$

where $W_O$ is a weight matrix and each head $i$ is computed as:

$$\text{head}_i = \text{Attention}(Q_i, K_i, V_i) \quad (5)$$

MLP is two fully connected networks in a feed-forward fashion. Residual connections are used between each layer of the TE as shown in Figure 1. Dropout is a regularisation technique designed to prevent overfitting during training.

## 3 Related works

The primary objective of this research is the identification of brain tumours. This section includes the latest methods applied to brain tumour identification approaches. Notable deep-learning solutions for accurately identifying brain tumours are included in this section.

### *3.1 Classification of brain tumours*

This subsection provides existing research in brain tumour classification.

Khan et al. (2022a) proposed an approach that used two deep-learning architectures to identify and classify multiclass brain tumours. They combined the VGG16 model with their proposed 23-layer CNN architecture. Transfer learning and data augmentation techniques were used to avoid overfitting. Sultan et al. (2019) proposed a 16-layered CNN model including convolutional layers, ReLU activation functions, normalisation, max-pooling layers, dropout layers, and fully connected layers. They used data augmentation techniques to expand the dataset and prevent overfitting. Noreen et al. (2020) proposed a hybrid architecture. This research used two different CNN models. Each model has different sizes of feature maps. In this research, two pre-trained models, Inception-v3 and DenseNet201 were used to extract multi-level features. The extracted features were concatenated from different layers of both models, providing more comprehensive information for classifying brain tumours. Kang et al. (2021) proposed an ensemble of deep-learning models. The study used 13 different pre-trained CNN models for the extraction of deep features. The top three deep features were concatenated to form a more robust feature representation. This ensemble of deep features was then fed into machine learning classifiers such as support vector machine (SVM), random forest (RF), and K-nearest neighbours (KNN) for final classification.

Khan et al. (2020) used two variants of the same CNN model, VGG16 and VGG19. Pre-trained versions of both models were used for feature extraction and those features were concatenated. Irmak (2021) proposed a CNN = based model for the multi-classification of brain tumours. The proposed models were compared with popular state-of-the-art CNNs (e.g., AlexNet, InceptionV3, ResNet-50, and VGG-16). Additionally, there are numerous research based on CNNs to classify the types of tumours accurately (Chauhan et al., 2017; Saleh et al., 2020; Sharif et al., 2021; Waghmare and Kolekar, 2021; Mohsen et al., 2018; Aamir et al., 2022; Paul et al., 2017; Deepak and Ameer, 2019). However, the classification of the tumours does not provide the exact location of the tumours in the MRI. The classification only includes information on whether the specific MRI contains tumours. The main problem is to approximate the location and types of the tumours.

### *3.2 Detection and segmentation of brain tumours*

This subsection provides notable research in brain tumour detection and segmentation. Detection and segmentation of the tumours help to locate and identify the tumour type. Mohsen et al. (2018) used fuzzy C-means clustering to segment the MRI images into different sections, focusing on isolating the tumour regions. This study applied the discrete wavelet transform (DWT) to extract features from the segmented images. Principal component analysis (PCA) was then used to reduce the dimensionality of the feature vectors. This research used a deep neural network (DNN) with seven hidden layers for classification. The performance of the DNN classifier was compared against other machine learning models like KNN, linear discriminant analysis (LDA) and SVM. This method relies on the segmentation step (using fuzzy C-means) to isolate the tumour regions, which might introduce a dependency on accurate segmentation. If segmentation is not precise, it can affect the feature extraction and, consequently, the classification results. While the study compares DNN with other traditional machine learning models, it does not explore other advanced deep learning architectures like CNN for a more comprehensive evaluation.

Saeedi et al. (2023) used two deep learning architectures: CNN and auto-encoder. This research used CNN architecture to extract the features and combine an encoder network with a CNN-based classifier. This study also compared six machine learning classifiers (KNN, random forest, SVM, logistic regression, stochastic gradient descent and multilayer perceptron) on the same MRI dataset. The KNN model had the highest classification accuracy of 86%, while MLP had the lowest at 28%. Liu et al. (2023) proposed the segmentation of the brain tumours. This study provided a comprehensive study of deep learning-based approaches for brain tumour segmentation, addressing challenges like data imbalance, location uncertainty, and morphological diversity of tumours. Many researchers focus on the detection and segmentation of tumours (Chegraoui et al., 2021; Telrandhe et al., 2015; Arunkumar et al., 2019; Aleid et al., 2023; Kang et al., 2023; Mercaldo et al., 2023; Dipu et al., 2021; Suresha et al., 2020; Sal¸cin et al., 2019; Bhanothu et al., 2020). However, they failed to provide a comparative study of various object detection methods that match the accuracy and speed in detecting brain tumours.

This highlights the opportunity to develop a method for accurately identifying tumours, including pinpointing their exact location and determining the specific type of tumour. This research compares the 11 most popular object detection methods for identifying brain tumours. In addition, a novel improved YOLOv8 is proposed for the first time in brain tumour identification. The proposed method outperformed the existing 11 object detection methods in accuracy. This research provides the best solution for medical doctors or experts to locate the area and the type of tumours in MRI images, showcasing excellence.

## 4 Methodology

A simple pipeline of the proposed methodology is shown in Figure 2. The MRI brain images are first pre-processed to meet the requirements of the improved YOLOv8 model. These pre-processed images are then used to train and validate the proposed YOLOv8 model. Ultimately, the model is employed to identify tumours in the images. This section will briefly discuss the proposed methodology for effectively identifying brain tumours.

**Figure 2** A high-level sketch of the proposed brain tumour identification method methodology (see online version for colours)

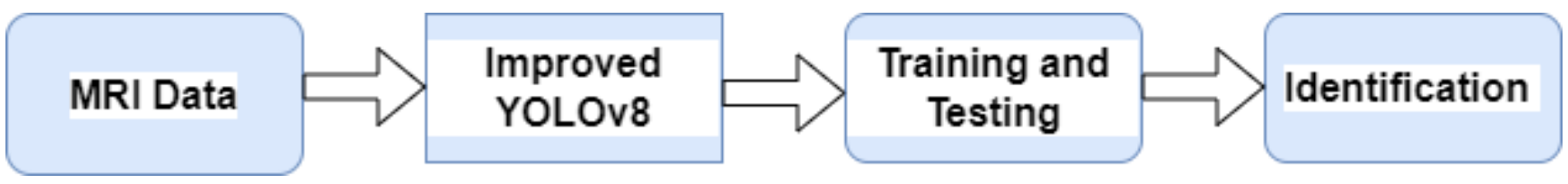

### *4.1 Dataset description*

We utilised a publicly available dataset from Kaggle (n.d.), which includes MRI images of brain tumours, specifically glioma and meningioma, and images without tumours. The dataset is divided into 878 images for training and 223 for validation, providing diverse examples to train and evaluate our model effectively.

### *4.2 Data pre-processing*

The dataset is annotated with bounding boxes using Roboflow (n.d.) to adhere to the YOLOv8 format. We applied data augmentation techniques to enhance the model's robustness and increase the number of training samples. These techniques include rotations of +15 and –15 degrees and horizontal and vertical flips. These augmentations help the model generalise better by exposing it to various image orientations and perspectives.

**Figure 3** The architecture of the YOLOv8 (see online version for colours)

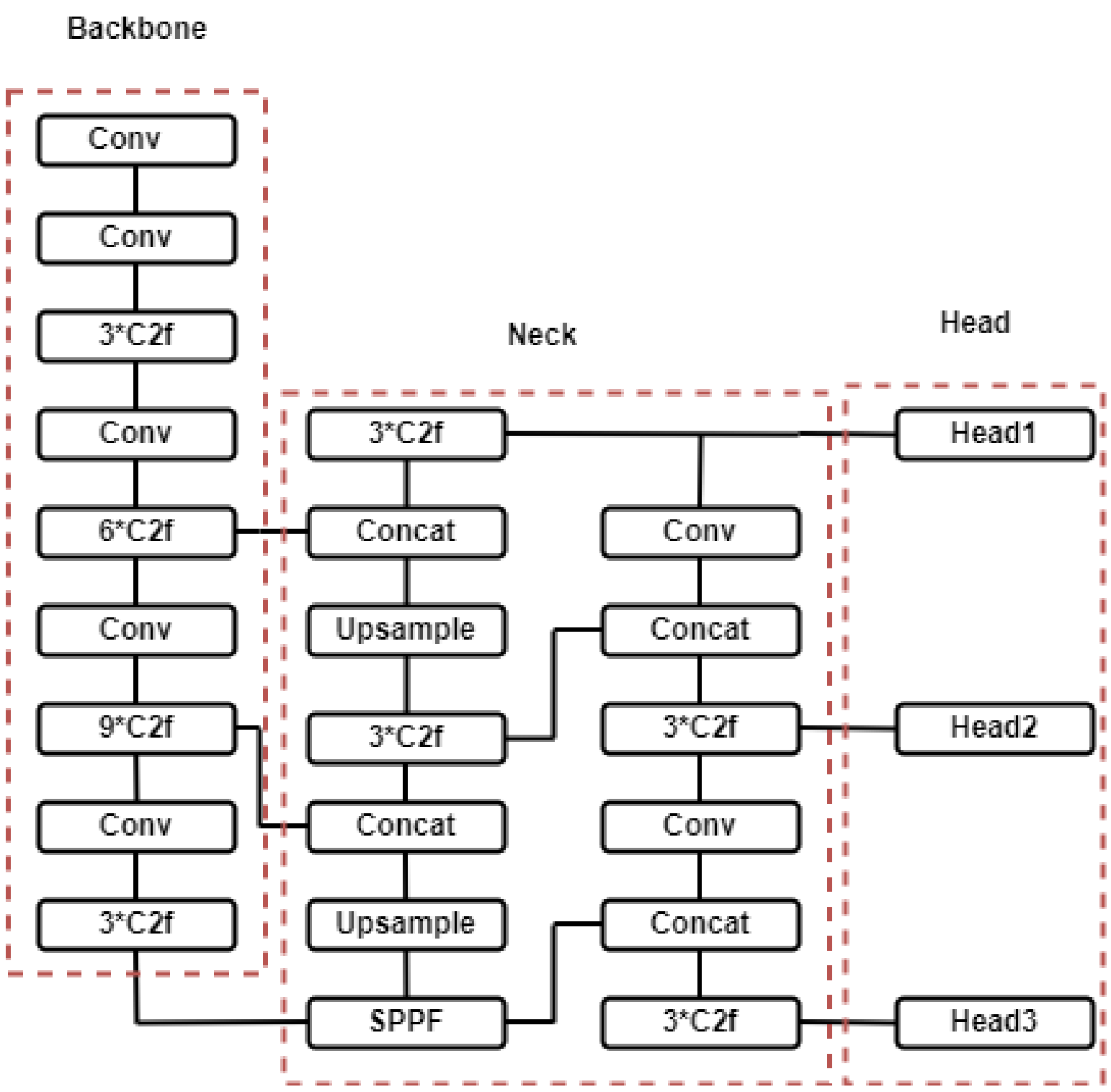

Note: It consists of a backbone, neck and head modules.

### *4.3 YOLOv8*

To fully understand the improved YOLOv8, we first describe the architecture of YOLOv8. YOLOv8 is the combination of various modules and techniques used in deep learning. The architecture of YOLOv8 (Ultralytics, n.d.) is sketched in Figure 3. YOLOv8 is a one-stage object detection model that consists of backbone, neck, and head modules. Backbone is used to extract the features from the images in different scales. The neck combines the different features and passes to the head. The head is used to detect (classify and localise) the object. A detailed explanation of each module is presented as follows:

### *4.3.1 Backbone*

The backbone comprises several convolution layers (Conv) blocks, a C2f module, and spatial pyramid pooling-fast (SPPF). Conv is a convolution block composed of a 2D convolution applied to the input image, batch normalisation (Ioffe and Szegedy, 2015), and sigmoid-weighted linear unit (SiLU) (Elfwing et al., 2018) activation function. SiLU is an improvement over rectified linear unit (ReLU) (Agarap, 2018). SiLU does not have fixed upper or lower bounds, unlike the sigmoid function (which is bounded between 0 and 1). This allows SiLU to avoid saturation issues where gradients can vanish, a common problem with sigmoid in deep networks.

The C2f module in YOLOv8 is a streamlined and faster variant of the CSP (Wang et al., 2020) Bottleneck with two convolutions (C2). Designed to enhance the execution speed without compromising performance. C2f introduces modifications that optimise the original C2 module. The C2f module consists of an initialisation phase where the input channels are mapped to fewer hidden channels using a convolutional layer. This initial layer splits the input into two parts, which are then processed in parallel. The module includes a series of bottleneck blocks, each processing the output from the preceding block. The processed outputs are then concatenated along with the initial split and passed through another convolutional layer, which combines and refines the features. An alternative forward method further optimises the processing, offering additional speed improvements. Overall, the C2f module is an efficient building block that maintains the architectural integrity of the original CSP bottleneck while providing faster processing for the YOLOv8 model. The different numbers (like 3 numbers of C2f represented by 3*C2f in Figure 3) of the C2f blocks are stacked by adopting residual connection, and feature maps extracted by three levels of C2f are forwarded to the neck module. These three levels (also called feature layers) of C2f have feature maps of different dimensions. The first layer from the top extracts less semantic information. As the depth of the backbone increases, more semantic information is extracted, and the size of feature maps keeps decreasing. Thus, the lowest feature layer extracts high semantic details, but the feature map size is the smallest.

### *4.3.2 Neck*

The outputs of the C2f blocks are passed to the SPPF block. It is an improvement of spatial pyramid pooling (SPP) (He et al., 2015) by replacing large-sized kernels with small-sized kernels for faster operations. SPPF is a powerful technique that provides a multi-scale representation of input feature maps, capturing features at various levels of abstraction (dimensions). This capability is beneficial for object detection, where detecting objects of different sizes is essential for achieving high accuracy and robustness.

The neck of YOLOv8 is inspired by the working mechanism of the feature pyramid network (FPN) (Lin et al., 2017a) and path aggregation network (PANet) (Liu et al., 2018). FPN is a pyramid-style structure mainly used to combine features of different scales. The backbone extracts features at three levels of abstraction with varying sizes. These different-sized feature maps are upsampled to match the size of feature maps of different feature levels and concatenated. PANet enhances the FPN architecture by incorporating a bottom-up path augmentation. This additional pathway allows position information from lower levels of the network to be effectively transmitted to higher levels. As a result, the positioning capabilities of the network are improved across multiple scales. This bottom-up approach complements the original top-down pathway of the backbone, ensuring that detailed spatial information from early layers is preserved and utilised in deeper layers, leading to better performance in tasks requiring multi-scale feature representation.

### *4.3.3 Head*

In YOLOv8, three heads are designed to detect objects of various sizes. Specifically, these heads target objects at three different scales. They are for large, medium, and small objects. Figure 3 shows three heads: head1 is designed to detect large objects, head2 for medium objects, and head3 for small objects. The heads create grids on these feature maps based on their respective dimensions. Each grid cell is responsible for predicting bounding boxes at its location. Three groups of anchors with different aspect ratios are predefined for each grid cell on each feature map. These anchors serve as reference bounding boxes to generate candidate bounding boxes for object detection. After generating the candidate bounding boxes, non-maximum suppression (NMS) (Neubeck and van Gool, 2006) is applied. NMS is a post-processing step that removes overlapping bounding boxes by keeping only the ones with the highest confidence scores. This step ensures that the final output consists of bounding boxes with their locations, sizes, and the associated confidence scores of the detected objects, minimising redundancy and improving detection accuracy. The head provides the bounding box of the detected objects, class name, and the confidence score.

## *4.4 Improved YOLOv8*

The proposed YOLOv8 is motivated by the success of ViT (Dosovitskiy et al., 2020) in computer vision tasks. ViT is a transformer (Vaswani et al., 2017)-based architecture adapted for computer vision tasks which shows state-of-the-art performance in image recognition (Dosovitskiy et al., 2020). It replaces the convolutional layers in traditional computer vision models with self-attention mechanisms. This allows the model to attend to different regions of an image flexibly and adaptively, extracting rich features.

The sketch of the proposed architecture is shown in Figure 4. The proposed model modifies the YOLOv8 model in the backbone, neck, and head modules by replacing the last C2f block of the backbone with the vision transformer encoder (TE) block and the Conv block with the GhostConv block. The YOLOv8 head has NMS as the post-processing step. But they are hand-designed and fixed components. They cannot be changed according to the size of the objects. This means that the same IoU threshold and suppression criteria are applied uniformly across all detected objects, regardless of their size or context. As a result, the fixed nature of NMS might not be optimal for all situations, mainly when dealing with objects of varying sizes or in cases where the standard IoU threshold does not suit the specific detection scenario (Carion et al., 2020). To address the limitations of the NMS, a dynamic head is used in this improved YOLOv8. The improved YOLOv8 has an RT-DETR head, providing more accurate identification accuracy without hand-designed NMS components.

**Figure 4** The proposed architecture of improved-YOLOv8 (see online version for colours)

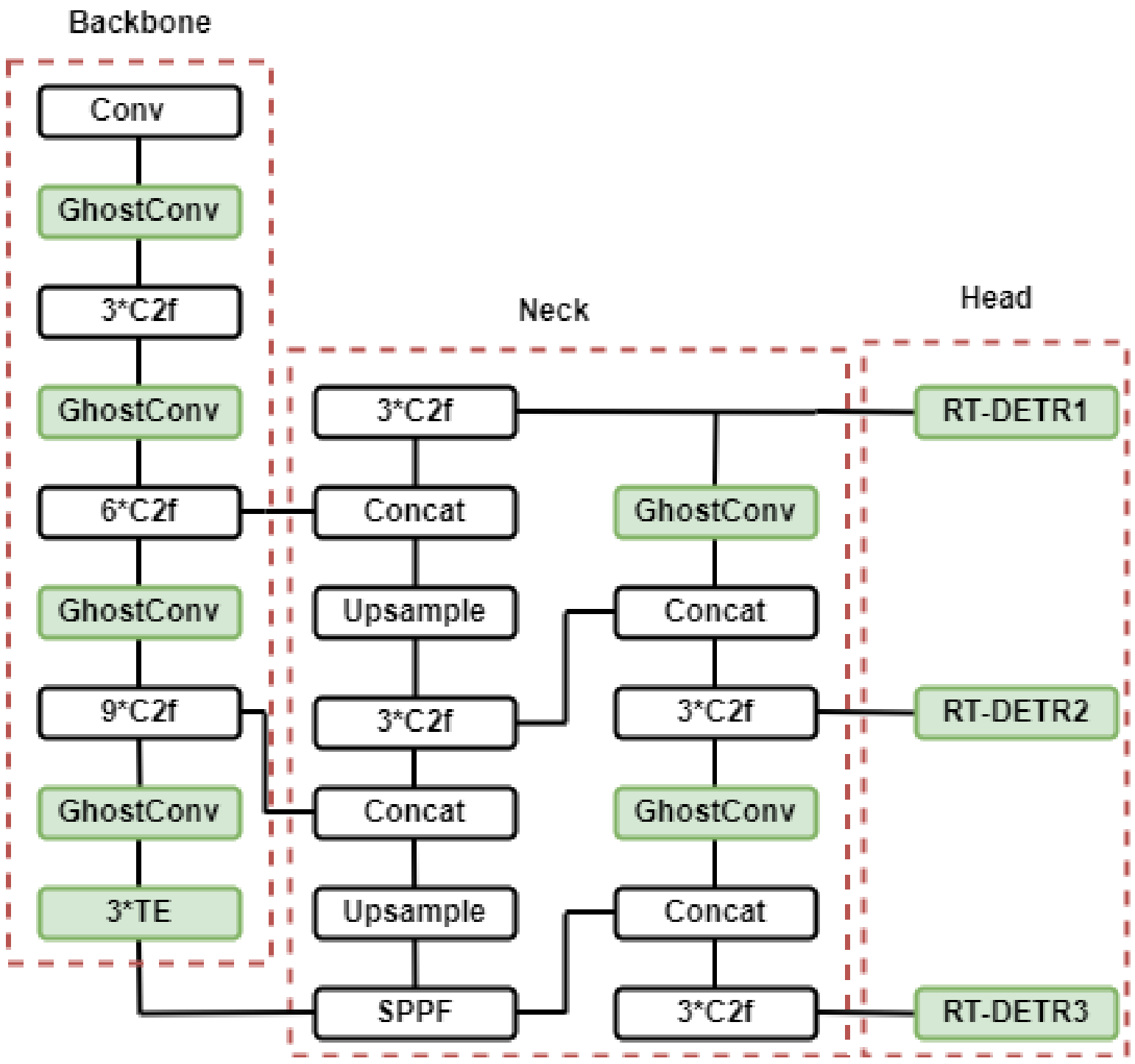

Note: It has modifications on the backbone, neck, and head modules. The green boxes represent the improved part.

### 4.5 *Implementation*

The experiment was conducted and implemented using the Pytorch library. For all experiments conducted for this research, the same hyperparameters are used to implement all selected models and are given in Table 1.

**Table 1** Hyperparameters for the conducted experiments in this research

| | |
|---|---|
| Epochs | 200 |
| Batch size | 16 |
| Learning rate | 0.01 |
| Momentum for SGD | 0.95 |
| Weight decay for regularisation | 0.0005 |
| Graphical processing unit (GPU) | Tesla T4, 15 GB |

### 4.6 *Evaluation criteria*

The performance of the selected models is evaluated using standard metrics, such as mAP. mAP is an essential evaluation metric for object detection models, offering insights into the models' accuracy and speed. mAP is a common criterion for measuring object detection performance. The mAP is calculated by calculating the average precision (AP) for each class and then the mean value of AP over all classes as below:

$$Precision\ (P) = \frac{TP}{TP + FP} \tag{6}$$

$$Recall\ (R) = \frac{TP}{TP + FN} \tag{7}$$

$$AP = \int_0^1 P(R), dR \tag{8}$$

where $P(R)$ denotes precision ~~as a functio~~n of recall, and $dR$ represents an infinitesimal change in recall. This integral effectively sums the precision values across different recall thresholds, capturing the model's overall performance. $TP$, $FP$, and $FN$ stand for true positive, false positive, and false negative, respectively.

- TP: A true positive is when the model correctly identifies and localises an object within an image. This means the predicted bounding box sufficiently overlaps with the ground truth bounding box of the object. Ground truth refers to the actual, real-world data manually labelled or annotated. It serves as a reference or benchmark against which the performance of models is evaluated.
- FP: A false positive is when the model detects an object not in the image or mislocalises the object by providing a bounding box that significantly deviates from the actual object location or ground truth.
- FN: A false negative in object detection refers to a situation where the model cannot detect an object in an image. This happens when the model either misses the object entirely or provides a bounding box that does not adequately capture the object.

In practice, AP is computed numerically by interpolating precision values at different recall levels and summing discrete areas under the curve.

$$mAP = \frac{1}{c}\sum_{i=1}^{c} AP_i \tag{9}$$

where $c$ is the classes number. The metric mAP@0.5 refers to the mAP computed when the IoU exceeds 50%. IoU measures the overlap between a predicted bounding box and the ground truth bounding box, providing a quantitative assessment of how well the predicted object aligns with the actual object. This score comprehensively assesses an object detection model's accuracy across various classes. mAP@0.5 is used as a threshold in the evaluation metric because it is a standard metric and threshold used by PASACL VOC object detection challenge (Everingham et al., 2015).

### *4.7 Comparison with existing object detection models*

The comparative analysis of the performance of the selected models and our proposed model is presented in Table 2.

**Table 2** Comparative result of the brain tumour identification performance

| *Models* | *mAP@0.5* |
|---|---|
| Faster R-CNN | 0.68 |
| Mask R-CNN | 0.72 |
| SSD | 0.61 |
| RetinaNet | 0.73 |
| EfficientDet | 0.82 |
| DETR | 0.79 |
| YOLO | 0.65 |
| YOLOv3 | 0.85 |
| YOLOv4 | 0.84 |
| YOLOv5 | 0.88 |
| YOLOv8 | 0.87 |
| *Improved YOLOv8 (our)* | *0.91* |

Table 2 presents a comparative analysis of various object detection models for brain tumour identification, showcasing their performance based on the mAP at a 0.5 IoU threshold. Traditional models like faster R-CNN and SSD show moderate performance with mAP scores of 0.68 and 0.61, respectively. More advanced models such as RetinaNet, EfficientDet, and DETR demonstrate improved accuracy, with mAP scores ranging from 0.73 to 0.82. The YOLO series of models, particularly YOLOv3, YOLOv4, and YOLOv5, achieve higher accuracy, with mAP scores of 0.85, 0.84, and 0.88, respectively. Notably, the improved YOLOv8 model proposed in our work outperforms all other models, achieving the highest mAP of 0.91, highlighting its superior effectiveness in accurately identifying brain tumours.

The mAP at IoU threshold 0.5 (mAP@0.5) has been plotted throughout the epochs to assess the performance of the models employed in this experiment. This metric comprehensively evaluates the models' accuracy in identifying the tumours, offering insights into their effectiveness across the training process. As shown in Figure 5, the plot allows a detailed comparison of model performance at different stages, highlighting improvements or stagnation throughout the training epochs. The x-axis represents the number of training epochs, while the y-axis represents the mAP@50 score. The proposed improved YOLOv8 model, represented by the red line, demonstrates superior performance, achieving a higher mAP@50 than all other models. It converges faster, reaching high accuracy within fewer epochs, and consistently performs throughout training. Compared to the standard YOLOv8, the proposed model exhibits a noticeable improvement, indicating the effectiveness of its modifications. Additionally, its stable mAP curve after convergence suggests better generalisation. These results highlight that the proposed model outperforms existing models in accuracy and training efficiency, making it a more effective tumour identification model.

Figure 6 presents a sample of tumours detected using the proposed model. The bounding boxes highlight the approximate location of the tumours, while the corresponding labels indicate the identified tumour types. This information provides valuable insights for medical professionals, enabling a more efficient and accurate assessment. By clearly delineating tumour regions and categorising them, the proposed model enhances the interpretability of medical images, potentially assisting in early diagnosis and treatment planning and improving overall clinical decision making.

**Figure 5** Performance evaluation of models across epochs: mAP@0.5 plot showing tumour identification accuracy throughout the training (see online version for colours)

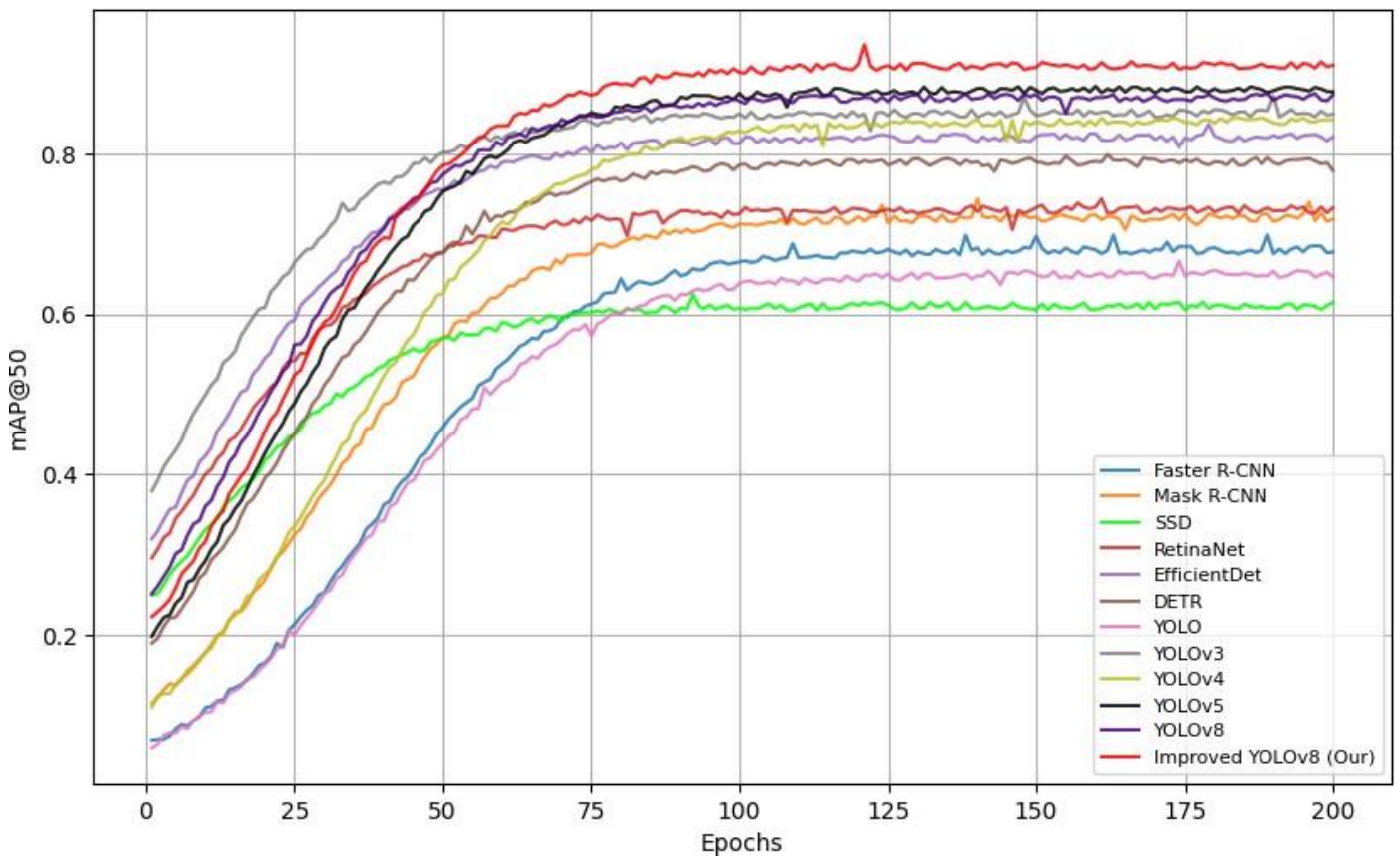

**Figure 6** A sample of the identified tumours using improved YOLOv8 model (see online version for colours)

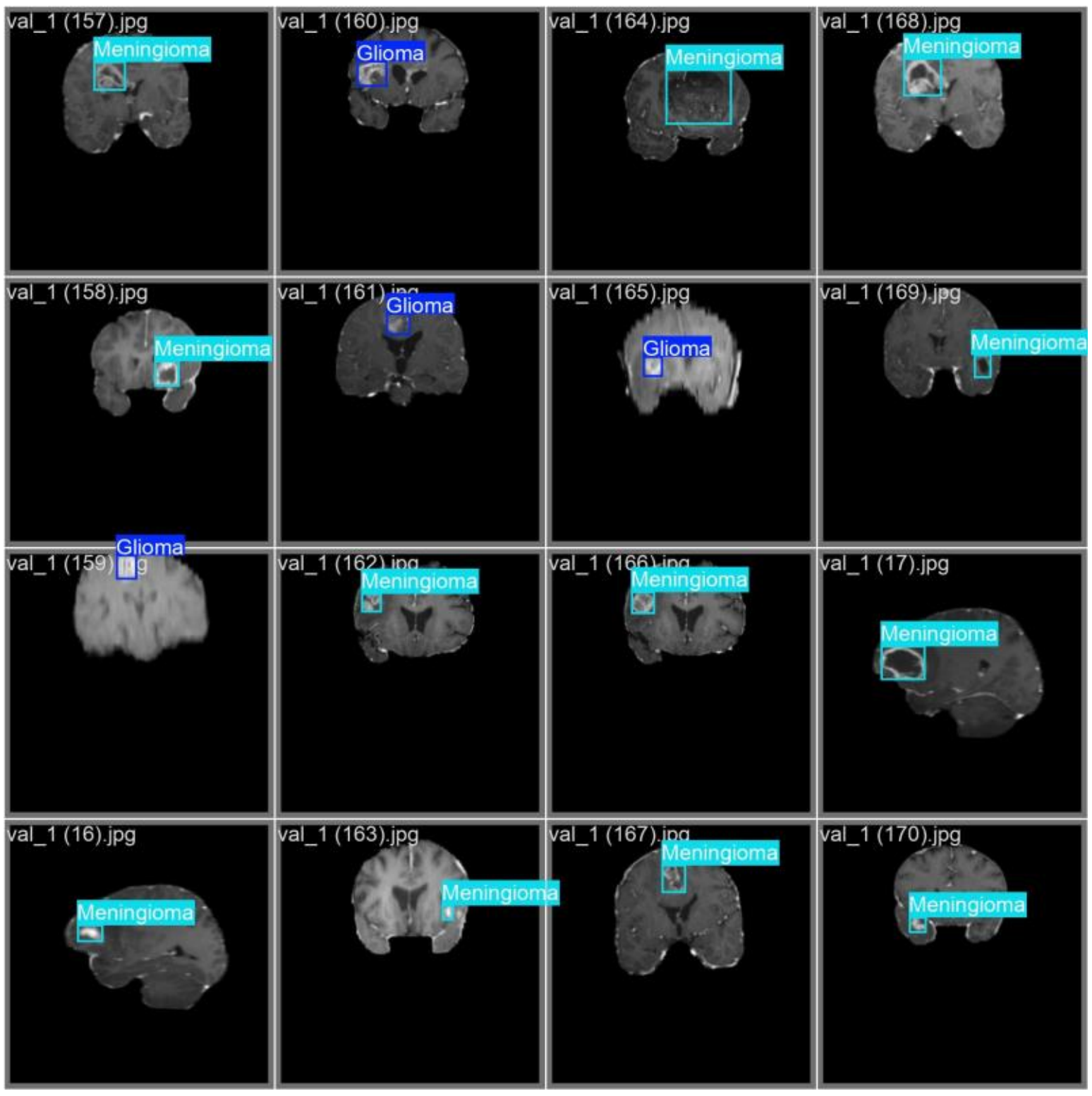

Note: The bounding box provides the location of the tumours and the type of tumours.

YOLOv5 has the second-highest best performance of 0.88 after the proposed model. Thus, we have shown the sample of the identified tumours by the YOLOv5 model in Figure 7. Comparing two identification results, as shown in Figures 6 and 7, the YOLOv5 leaves unidentified tumours for the image val_1(161.jpg). The ground truth label of the image val_1(161.jpg) is glioma, which the improved YOLOv8 model correctly identifies. Similarly, the image val_1(106.jpg) is misidentified. This image is identified as two types of tumours by YOLOv5. The ground truth label is meningioma, which the improved YOLOv8 model correctly identifies.

**Figure 7** A sample of the identified tumours using YOLOv5 (see online version for colours)

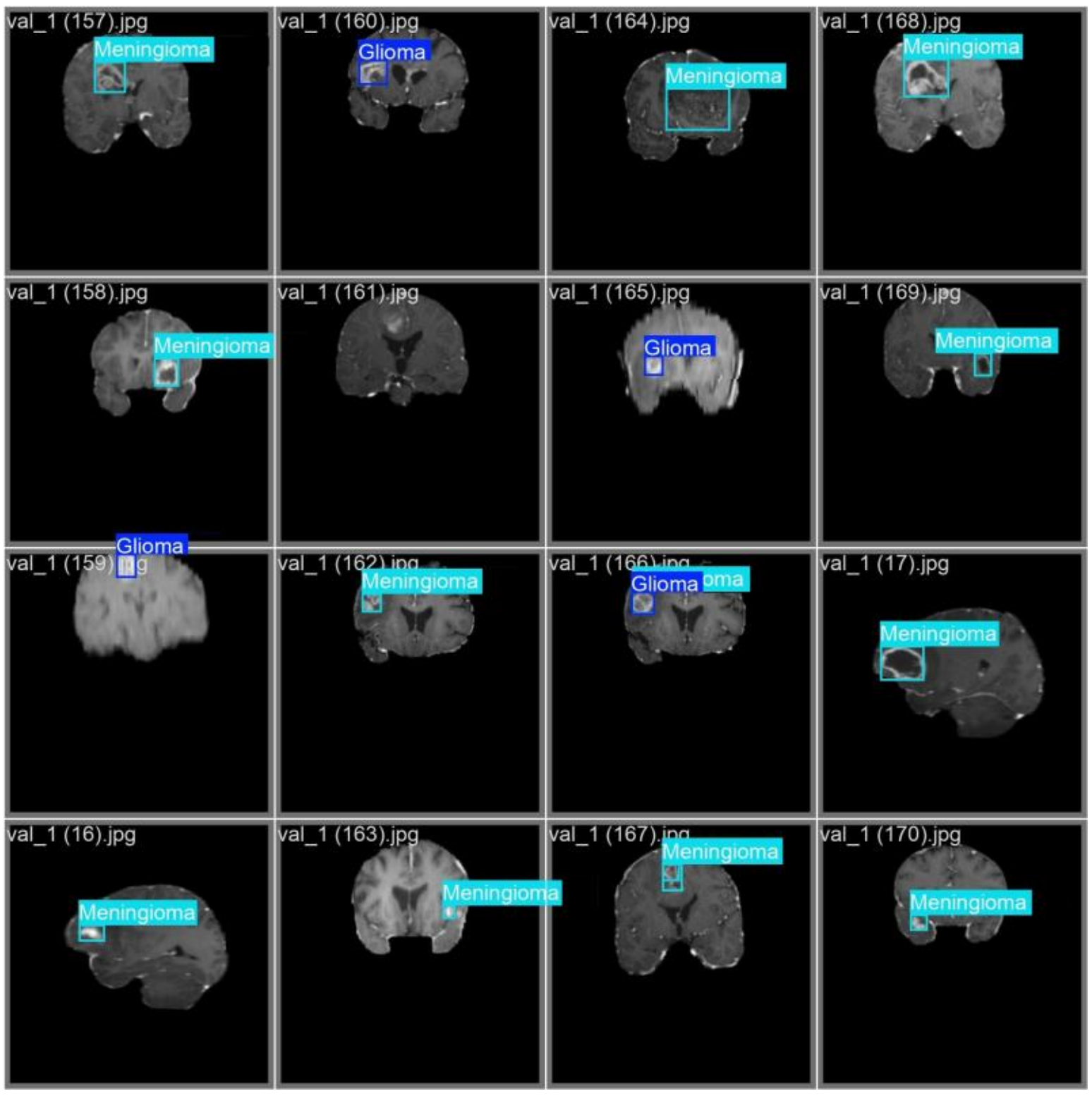

Note: This figure shows some images remain unidentified, and some are misidentified (as both types of tumours).

The improved YOLOv8 proposed in this research shows impressive performance for the following reasons.

- Replacing the last C2f block with a TE block introduces better global context understanding. Transformers excel at capturing long-range dependencies and relationships between different parts of an image, which can enhance feature representation, especially in complex scenes where contextual information is crucial. Integration of the TE block provides high-quality feature extraction and extra attention or focus on crucial areas of brain tumour images to identify them correctly and precisely.

- Ghost convolution is designed to reduce the number of parameters and computational costs while maintaining or even improving the representative capacity of the model. Generating 'ghost' feature maps that approximate the full convolutional operation makes the model more efficient and lightweight. The integration of ghost convolution decreased the parameters from 11.2 to 9.04 million. This reduction in parameters makes the model light and requires less computational power.
- Replacing the standard YOLOv8 head with RT-DETR improves the detection process by leveraging the strengths of transformer-based detection heads. RT-DETR is designed for real-time object detection, offering efficient handling of object relations and better performance, particularly in complex scenes. It also typically has more adaptive post-processing, which could address some limitations of fixed NMS providing dynamic tumour identification.
- The combination of ViT and ghost convolution improves feature extraction at multiple levels, ensuring that the model captures local and global features more effectively. This results in more robust representations, leading to better detection accuracy.

## 5 Conclusions

In conclusion, our research presents a significant advancement in the automated detection of brain tumours by introducing an enhanced YOLOv8 model. Through strategic modifications, including integrating a ViT block, ghost convolution, and RT-DETR, our proposed model achieves a final accuracy of 0.91 mAP@0.5. This performance surpasses 11 popular object detection methods validated on a publicly available dataset. Our model's improved accuracy and efficiency provide a reliable tool for medical professionals, aiding in the accurate and timely detection of brain tumours using MRI images. This contribution can potentially enhance diagnostic processes and patient outcomes in clinical settings.

## Declarations

All authors declare that they have no conflicts of interest.